%% file: main.tex
\documentclass[twoside]{article}

\usepackage[accepted]{aistats2021}
\usepackage{color}
\definecolor{cgreen}{rgb}{0.2,0.6,0.2}
\definecolor{darkred}{rgb}{0.4,0.0,0.0}
\definecolor{darkgreen}{rgb}{0.0,0.4,0.0}
\definecolor{darkblue}{rgb}{0.0,0.0,0.4}
\usepackage[colorlinks=true,
            linkcolor=red,
            urlcolor=blue,
            citecolor=darkblue]{hyperref}
            
\usepackage{xspace}
\usepackage{url}
\usepackage{amsmath}
\usepackage{amsthm}
\usepackage{amssymb}
\usepackage{mathtools}
\usepackage{multirow}
\usepackage{titlesec}
\usepackage{csquotes}
\usepackage{graphicx}
\usepackage{subcaption}
\usepackage{tikz}
\usepackage{bbm}
\usepackage{physics}
\usepackage{verbatim}
\usepackage{dsfont}
\usepackage{cleveref}
\usepackage{stmaryrd,scalerel}
\usepackage[ruled,vlined]{algorithm2e}
\usepackage{booktabs}
\usetikzlibrary{arrows}
\usetikzlibrary{positioning}

\newtheorem{remark}{Remark}

\newcommand{\thetav}{\boldsymbol{\theta}}

\newcommand{\Lc}{\mathcal{L}}

\newcommand{\Usf}{\mathsf{U}}

\newcommand{\Fc}{\mathcal{F}}
\newcommand{\KL}{\mathsf{KL}}
\newcommand{\Zsf}{\mathsf{Z}}
\newcommand{\Xsf}{\mathsf{X}}

\newcommand{\xv}{\mathbf{x}}
\newcommand{\rmd}{\mathrm{d}}
\newcommand{\gB}{\mathcal{B}}
\newcommand{\zv}{\mathbf{z}}
\newcommand{\yv}{\mathbf{y}}

\newcommand{\Eb}{\mathbb{E}}

\newcommand{\RR}{\mathds{R}}

\newcommand{\uv}{\mathbf{u}}

\newcommand{\Ib}{\mathbb{I}}
\newcommand{\Nc}{\mathcal{N}}

\newcommand{\eg}{{e.g.}\xspace}

\newcommand{\ie}{{i.e.}\xspace}
\newcommand{\iid}{{i.i.d.}\xspace}
\newcommand{\cf}{{cf.}\xspace}
\newcommand{\wrt}{{w.r.t.}\xspace}

\newcommand{\Tb}{\mathbf{T}}
\newcommand{\Qb}{\mathbf{Q}}

\newif\ifrev

\newcommand{\varv}{\boldsymbol{\varepsilon}}
\newcommand{\mo}{\mathsf{model}}
\usepackage[round]{natbib}
\renewcommand{\bibname}{References}
\renewcommand{\bibsection}{\subsubsection*{\bibname}}

\begin{document}

\runningtitle{Sampling in Combinatorial Spaces with SurVAE Flow Augmented MCMC}


\twocolumn[

\aistatstitle{Sampling in Combinatorial Spaces with \\ SurVAE Flow Augmented MCMC}

\aistatsauthor{Priyank Jaini \And Didrik Nielsen \And  Max Welling }

\aistatsaddress{Bosch-Delta Lab\\ University of Amsterdam \And Technical University of Denmark \And Bosch-Delta Lab \\ University of Amsterdam } ]

\begin{abstract}
Hybrid Monte Carlo is a powerful Markov Chain Monte Carlo method for sampling from complex continuous distributions. However, a major limitation of HMC is its inability to be applied to discrete domains due to the lack of gradient signal. In this work, we introduce a new approach based on augmenting Monte Carlo methods with SurVAE Flows to sample from discrete distributions using a combination of  neural transport methods like normalizing flows and variational dequantization, and the Metropolis-Hastings rule. Our method first learns a continuous embedding of the discrete space using a surjective map and subsequently learns a bijective transformation from the continuous space to an approximately Gaussian distributed latent variable. Sampling proceeds by simulating MCMC chains in the latent space and mapping these samples to the target discrete space via the learned transformations. We demonstrate the efficacy of our algorithm on a range  of  examples  from  statistics,  computational  physics and machine  learning, and observe improvements compared to alternative algorithms.
\end{abstract}
\input{sections/intro}
\input{sections/prelim_main}

\input{sections/method}
\input{sections/prev}
\input{sections/exp}

\input{sections/con}
%\newpage
\bibliography{bqn}
\bibliographystyle{apalike}

\newpage
\onecolumn
\appendix
\input{sections/appendix}

\end{document}

%% file: sections/intro.tex
\section{Introduction}
\label{sec:intro}

The ability to draw samples from a known distribution is a fundamental computational challenge. It has applications in diverse fields like statistics, probability, and stochastic modeling where  these  methods  are  useful  for  both estimation and inference. These are further useful within the frequentist inference framework to  form confidence  intervals  for  a  point  estimate.  Sampling procedures are also standard in the Bayesian setting for exploring posterior distributions, obtaining credible intervals, and solving inverse problems. The workhorse algorithms in these settings are simulation-based, amongst which the Markov Chain Monte Carlo \cite{brooks2011handbook} method is the most broadly used method. Impressive advances have been made -- both in increasing efficiency and reducing computation costs -- for sampling using Monte Carlo methods over the past century. However, problems in discrete domains still lack an efficient general-purpose sampler.  

In this paper, we consider the problem of sampling from a known discrete distribution. Inspired by the recent success of deep generative models -- particularly neural transport methods like normalizing flows \citep{TabakVE10,TabakTurner13,RezendeMohamed15}-- in unsupervised learning, we propose a new approach to design Monte Carlo methods to sample from a discrete distribution based on augmenting MCMC with a \emph{transport map}. Informally, let $\pi(\thetav)$ be a discrete target distribution and $p(\zv)$ be a simple source density from which it is easy to generate independent samples \eg a Gaussian distribution. Then, a transport map $\Tb$ from $p(\zv)$ to $\pi(\thetav)$ is such that if $\zv_i \sim p(\zv)$ then $\Tb(\zv_i) \sim \pi(\thetav)$. 

The significance of having such a transport map $\Tb$ is particularly consequential: firstly, given such a map $\Tb$, we can generate samples from the target $\pi(\thetav)$. Secondly, these samples can be generated cheaply irrespective of the cost of evaluating $\pi(\thetav)$. Importantly, $\Tb$ affords us the ability to sample from the marginal and conditional distributions of $\pi(\thetav)$ using $p(\zv)$ given an appropriate structure. Indeed, this idea of using neural transport maps based on normalizing flows have been explored by \cite{ParnoMarzouk18} for continuous densities by learning a diffeomorphic transformation $\Tb$ from the source density $p$ to the target density $\pi(\thetav)$ where $\thetav \in \RR^d$. 

In this paper, we extend this to discrete domains using the recently proposed SurVAE Flow framework \citep{nielsen2020survae}. We first learn a transport map from the discrete space to a continuous space using a surjective transformation $\Qb$. However, such a continuous space is often highly multi-modal with unfavorable geometry for fast mixing of MCMC algorithms. Thus, we learn an additional normalizing flow $\Tb_{\phi}$ that transforms this \emph{complex} continuous space to a simple latent space with density $p(\zv)$ where sampling is easy. We finally sample from the desired target distribution by generating samples from the latent space and using the learned transformations $\Qb$ and $\Tb$ to \textit{push-forward} these samples to the target space. Our complete implementation allows parallelization over multiple GPUs, improving efficiency and reducing computation time.

The main contributions of this paper are:
\begin{itemize}
    \item We present a SurVAE augmented general purpose MCMC solver for combinatorial spaces. 
    \item We propose a new learning objective compared to previous methods in normalizing flows to train a transport map that ``Gaussianizes'' the discrete target distribution.
    %\item By leveraging the advances in flow-based models, our method is better suited than the alternatives for sampling in high dimensions. 
    %\item Our complete implementation allows parallelization over multiple GPUs, improving efficiency and reducing computation time. We will make the code for our sampler publicly available.
\end{itemize}

The rest of the manuscript is organized as follows: We begin in \S\ref{sec:bkg} by presenting a brief background on normalizing flows and setting up our main problem. In \S\ref{sec:method}, we provide details of our method which consists of two parts: learning the transport map and, generating samples from the target distribution. Subsequently, in \S\ref{sec:prev}, we put our present work in perspective to other approaches to sampling in discrete spaces. Finally, we perform empirical evaluation on a diverse suite of problems in \S\ref{sec:exp} to demonstrate the efficacy of our method. Our implementation is available at \url{https://github.com/priyankjaini/discFlowMH}

%% file: sections/prelim_main.tex
\section{Preliminaries and Setup}
\label{sec:bkg}

In this section, we setup our main problem, provide key definitions and notations, and formulate the idea for an MCMC sampler augmented with normalizing flows for sampling in discrete spaces.

Let $p, q$ be two probability density functions (\wrt the Lebesgue measure) over the source domain $\Zsf\subseteq \RR^d$ and the target domain $\Xsf \subseteq\RR^d$, respectively. Normalizing flows learn a \emph{diffeomorphic} transformation $\Tb : \Zsf \to \Xsf$ that allows to represent $q$ using $p$ via the change of variables formula \citep{Rudin87}:
\begin{align*}
q(\xv) &= p(\zv) / |\Tb'(\zv)| \\
&= p(\Tb^{-1}\xv) / |\Tb'(\Tb^{-1}\xv)|,
\end{align*}
where $|\Tb'(\zv)|$ is the (absolute value) of the Jacobian (determinant of the derivative) of $\Tb$. In other words, we can obtain a new random variable $\xv \sim q$ by pushing the source random variable $\zv \sim p$ through the map $\Tb$. When we only have access to an \iid sample $\lbag \xv_1, \ldots, \xv_n \rbag\sim q$, we can learn $\Tb$ and thus $q$ through maximum likelihood estimation: 
\begin{align*}
\max_{\Tb \in \Fc}~~ \frac{1}{n}\sum_{i=1}^n \Big[-\log |\Tb'(\Tb^{-1}\xv_i)| + \log p(\Tb^{-1}\xv_i) \Big].
\end{align*}
where $\Fc$ is a class of diffeomorphic mappings. Conveniently, we can choose \emph{any} source density $p$ to facilitate estimation \eg standard normal density on $\Zsf = \RR^d$ (with zero mean and identity covariance) or uniform density over the cube $\Zsf = [0,1]^d$.

This ``push-forward'' idea has played an important role in optimal transport theory \citep{Villani08} and has been used successfully for Monte carlo simulations. For example  \cite{MarzoukMPS16,ParnoMarzouk18,PeherstorferMarzouk18, hoffman2019neutra, albergo2019flow} have used normalizing flows for continuous random variables to address the limitations of HMC which suffers from the chain to mix slowly between distant states when the geometry of the target density is unfavourable.  

Specifically, \cite{ParnoMarzouk18} addressed this problem of sampling in a space with difficult geometry by learning a diffeomorphic transport map that transforms the original random variable to another random variable with a simple distribution. Concretely, let our interest be to sample from $\pi(\thetav)$ where $\thetav \in \Theta \subseteq \RR^d$. We can proceed by learning a diffeomorphic map $\Tb : \Zsf \to \Theta$ such that $\tilde{p}(\zv) = \pi(\thetav)\cdot |\Tb'(\zv)|$ where $\zv = \Tb^{-1}(\thetav)$ such that $p(\zv)$ has a simple geometry amenable to efficient MCMC sampling. Thus, samples can be generated from $\pi(\thetav)$ by running MCMC chain in the z-space and pushing these samples onto the $\Theta$-space using $\Tb$. The transformation $\Tb$ can be learned by minimizing the KL-divergence between a fixed distribution with simple geometry in the z-space \eg a standard Gaussian and $\tilde{p}(\zv)$ above. The learning phase attempts to ensure that the distribution $\tilde{p}(\zv)$ is approximately close to the fixed distribution with easy geometry so that MCMC sampling is efficient. A limitation of these works that use diffeomorphic transformations augmented samplers is that they are restricted to continuous random variables. This is primarily because the flow models used in these works can only learn density functions over continuous variables. 

\cite{uria2013rnade} introduced the concept of \emph{dequantization} to extend normalizing flows to discrete random variables. They consider the problem of estimating the discrete distribution $\pi(\thetav)$ given samples $\lbag \thetav_1, \thetav_2, \cdots, \thetav_n \rbag \sim \pi$ by ``lifting'' the discrete space $\Theta$ to a continuous one $\Xsf$ by filling the gaps in the discrete space with a uniform distribution \ie $\xv \in \Xsf$ is such that $\xv := \thetav + \uv|\thetav$ where $\thetav \in \Theta$ and $\uv|\thetav \sim \mathsf{Uniform}(0, 1)$. Subsequently, they learn the continuous distribution $p_{\mo}(\xv)$ over $\Xsf$ using a normalizing flow by maximizing the log-likelihood of the continuous model $p_{\mo}(\xv)$. \cite{theis2015note} showed that maximizing likelihood of the continuous model is equivalent to maximizing a lower bound on the log-likelihood for a certain discrete model $\pi_{\mo} := \int_{\uv} p_{\mo}(\thetav + \uv) \rmd \uv$ on the original discrete data. Thus, this learning procedure cannot lead to the continuous model degenerately collapsing onto the discrete data, because its objective is bounded above by the log-likelihood of a discrete model. \cite{ho2019flow++} extended the idea of uniform dequantization to propose \emph{variational dequantization} where in $\uv|\thetav \sim q(\uv|\thetav)$ instead of the uniform distribution. They learn the dequantizing distribution, $q(\uv|\thetav)$ from data by treating it as a variational distribution.  They subsequently estimate $\pi(\thetav)$ by optimizing the following lower bound:
\begin{align}
\label{eq:discpi}
    \Eb_{\thetav \sim \pi} [\log \pi_{\mo}(\thetav)] \geq \Eb_{\substack{\thetav \sim \pi\\ \uv \sim q(\uv | \thetav)}} \Bigg[ \log \frac{p_{\mo}(\xv)}{q(\uv | \thetav)}\Bigg]
\end{align}

%associated with $p_{\mo}(\xv)$ \ie $\pi_{\mo} := \int_{\uv} p_{\mo}(\thetav + \uv) \rmd \uv$. 

%\cite{ho2019flow++} extended normalizing flows to learn a discrete distribution $\pi(\thetav)$ given samples $\lbag \thetav_1, \thetav_2, \cdots, \thetav_n \rbag \sim \pi$ by proposing the idea of \emph{variational dequantization}. Here a continuous density model $p_{\mo}(\xv)$ of $\pi(\thetav)$ is learned -- where $\xv := \thetav + \uv$,  $\uv  \in \Usf \subseteq [0, 1)^d$, $\uv \sim q(\uv|\thetav)$ and $q(\uv|\thetav)$ is a dequantization distribution -- by maximizing the log-likelihood of the discrete model, $\pi_{\mo}$, associated with $p_{\mo}$ \ie  $\pi_{\mo} := \int_{\uv} p_{\mo}(\thetav + \uv) \rmd \uv$. By treating $q(\uv|\thetav)$ as an approximate posterior, they obtain the following bound:\MW{It is unclear if this is still work from \cite{ho2019flow++} you are describing or whether this is new.} \PJ{Yes, this is still from \cite{ho2019flow++}. Have rephrased the sentence to make it clear.}
%wherein $p_{\mo}(\xv)$ and $q(\uv|\thetav)$ are modelled using separate normalizing flows such that we have:
%\begin{align*}
    %&\Tb_{\lambda, \thetav} : \varv \to \Usf | \Theta, \qquad q(\uv|\thetav) := p_{\varv}(\varv)\cdot \Bigg| \frac{\partial \Tb_{\lambda,\thetav}(\varv)}{\partial \varv}\Bigg|^{-1} \\
   %\text{\&} \quad &\Tb_{\phi} : \Zsf \to \Xsf,\qquad p_{\mo}(\xv) := p_{\zv}(\zv)\cdot \Bigg|\frac{\partial \Tb_{\phi}(\zv)}{\partial \zv}\Bigg|^{-1} 
%\end{align*}
Recently, \cite{nielsen2020survae} introduced SurVAE Flows that extends the framework of normalizing flows to include surjective and stochastic transformations for probabilistic modelling. In the SurVAE flow framework, dequantization can be seen as a \emph{rounding} surjective map $\Qb_{\lambda} : \Xsf \to \Theta$ with parameters $\lambda$ such that $\thetav := \lfloor \xv \rfloor$ where the forward transformation is a discrete surjection $P(\thetav|\xv) = \Ib(\xv \in \gB(\thetav))$, for $\gB(\thetav) = \{\thetav+\uv|\uv\in[0,1)^d\}$. The inverse transformation $\Qb_{\lambda}^{-1} := q(\xv|\thetav)$ is stochastic with support in $\gB(\thetav)$. Thus, these ideas of using dequantization to learn a discrete probability distribution can be viewed as learning a transformation $\Qb_{\lambda}^{-1}$ that transforms a discrete space $\Theta$ to a continuous space $\Xsf$. %Since such a continuous embedding may have a complicated multi-modal geometry, a subsequent flow $\Tb$ transforms $\Xsf$ to $\Zsf$ with a simple geometry.

In this paper, we study the problem of sampling in discrete spaces \ie let $\Theta \subseteq \{1, 2, \cdots, K\}^d := [K]^d$ be a discrete random variable in $d$-dimensions with probability mass function (potentially unnormalized) $\pi(\thetav), ~\thetav \in \Theta$. Given access to a function that can compute $\pi(\thetav), ~\forall \thetav \in \Theta$, we aim to generate samples $\lbag \thetav_1, \thetav_2, \cdots, \thetav_n\rbag \sim \pi(\thetav)$. The aforementioned works on normalizing flows for discrete data is not directly applicable in this regime. This is due to the fact that \cite{uria2013rnade, ho2019flow++} and \cite{nielsen2020survae} estimate $\pi(\thetav)$ given samples from the original distribution by maximizing the log-likelihood of a discrete model. However, we only have access to the function $\pi(\thetav)$ and the learning method given in \Cref{eq:discpi} cannot be used in our setting. In the next section, we address this by extending the ideas of neural transport MCMC \citep{ParnoMarzouk18, hoffman2019neutra, albergo2019flow} and leveraging SurVAE Flows \citep{nielsen2020survae}.

%% file: sections/method.tex
\section{Flow Augmented MCMC}
\label{sec:method}

We discussed in \Cref{sec:bkg} the utility of normalizing flows augmented Monte Carlo samplers for overcoming the difficulties posed by unfavourable geometry in the target continuous distribution, as evidenced by the works of \cite{ParnoMarzouk18, hoffman2019neutra} and \cite{albergo2019flow}. We will now introduce our method of SurVAE Flow augmented MCMC for sampling in combinatorial spaces.

Informally, our method proceeds as follows: We first define a rounding surjective transformation $\Qb_{\lambda} : \Xsf \to \Theta$ with parameters $\lambda$ such that $\Xsf$ is a continuous embedding of the $\Theta$ space with density $q(\xv)$. Since the continuous embedding may be highly multi-modal with potentially unfavourable geometry for efficient MCMC sampling, we define an additional diffeomorphic transformation, $\Tb_{\phi} : \Zsf \to \Xsf$ with parameters $\phi$, from a simple latent space $\Zsf$ to $\Xsf$. Subsequently, we learn these transformations via maximum likelihood estimation. This concludes the learning phase of our algorithm. Finally, we generate samples from $\pi(\thetav)$ by running MCMC chains to generate samples from the learned distribution over $\Zsf$ and pushing-forward these samples to the $\Theta$ space using $\Qb_{\lambda}$ and $\Tb_{\phi}$. We elaborate each of these steps below.

$\Qb_{\lambda}$ is a surjective transformation that takes as input a real-valued vector $\xv \in \Xsf \subseteq \RR^d$ and returns the \emph{rounded} value of $\xv$ \ie $\thetav = \lfloor \xv \rfloor$. Thus, the forward transformation from $\Xsf$ to $\Theta$ is deterministic. The inverse transformation from $\Theta$ to $\Xsf$ is however stochastic since the random variable $\Xsf := \Theta + \Usf_{|\thetav}$ where $\Usf_{|\thetav} \subseteq [0, 1)^d$ is given by $\xv:= \thetav + \uv$ where $\uv \sim q(\uv|\thetav)$. Given access to $q(\uv|\thetav)$, we can evaluate the density $q(\xv)$ at a point $\xv \in \RR^d$ exactly as:
\begin{align}
    \label{eq:qx}
    q(\xv) = q(\thetav + \uv) := \pi(\thetav)\cdot q(\uv|\thetav)
\end{align}
since $\xv \in \mathcal{B}(\thetav)$. Thus, we need to learn $q(\uv|\thetav)$ in-order to fully specify this surjective transformation. Since, $q(\uv|\thetav)$ can be any arbitrary continuous density, we learn $q(\uv|\thetav)$ using a normalizing flow \ie we learn a diffeomorphic transformation $\Tb_{\lambda}(\cdot ; \thetav) : \mathcal{E} \to \Usf_{|\thetav}$ where $\mathcal{E} \subseteq \RR^d$ is standard Gaussian distributed. Under this setup, we get \begin{align}
    \label{eq:dequan}
    q(\uv|\thetav) = p(\varv)\cdot |\Tb_{\lambda}'(\varv ; \thetav)|^{-1}
\end{align}
Thus, learning the surjective transformation $\Qb$ is equivalent to learning $q(\uv|\thetav)$ which reduces to learning a flow $\Tb_{\lambda}(\cdot; \thetav)$. Thus, for brevity we use the informal notation that the transformation $\Tb_{\lambda}(\cdot; \thetav)$ quantizes the continuous space $\Xsf$ to $\thetav$. Finally, using \Cref{eq:dequan}, the density $q(\xv)$ in \Cref{eq:qx} for a fixed $\Tb_{\lambda}(\cdot; \thetav)$ can be written as:
\begin{align}
    \label{eq:varQ}
    q(\xv) = \pi(\thetav)\cdot p(\varv)\cdot|\Tb_{\lambda}'(\varv ; \thetav)|^{-1}
\end{align}

The density $q(\xv)$ over the continuous embedding $\Xsf$ of $\Theta$ learned above can have any arbitrary geometry and may not be efficient for MCMC sampling. Thus, next we learn a diffeomorphic transformation $\Tb_{\phi}$ from a latent space $\Zsf$ to $\Xsf$ such that using the change of variables formula we get:%\MW{Is the next equation correct? It haven't seen it above in the general text} \PJ{This is change of variables with the inverse transformation. Can be obtained directly from the first equation in section 2. Note we now want p(z) in terms of q(x). The original eq. was missing an inverse though. Made that change.}
\begin{align}
    \label{eq:pztilde}
    \tilde{p}_{\phi, \lambda}(\zv) = q(\xv)\cdot|\Tb_{\phi}'(\Tb^{-1}_{\phi}\xv)|
\end{align}

Our complete model, therefore, consists of transformations $\Tb_{\phi}$ and $\Tb_{\lambda}(\cdot; \thetav)$ that transforms $\Zsf$ to $\Theta$. The next challenge is to learn $\Tb_{\phi}$ and $\Tb_{\lambda}(\cdot; \thetav)$ such that the induced density $\tilde{p}_{\phi, \lambda}(\zv)$ has a simple geometry for efficient MCMC sampling. We achieve this by forcing $\tilde{p}_{\phi, \lambda}(\zv)$ to be close to a standard Gaussian by minimizing the KL-divergence between $\tilde{p}_{\phi, \lambda}(\zv)$ and $p(\zv)$ where $p(\zv) = (2\pi)^{-\frac{d}{2}} \mathsf{exp}(-\zv^T\zv)$. 

Concretely, let $\Lc(\lambda, \phi) := \KL\big(p(\zv)~ || ~\tilde{p}_{\phi, \lambda}(\zv)\big)$. $\Lc(\lambda, \phi)$ can be approximated with the empirical average by generating \iid samples $\lbag \zv_1, \zv_2, \cdots, \zv_m \rbag \sim p(\zv)$ giving:
\begin{align*}
    \Lc(\lambda, \phi) \approx \frac{1}{m} \sum_{i=1}^m \log \frac{ p(\zv_i)}{ \tilde{p}_{\phi, \lambda}(\zv_i)}
\end{align*}
We thus arrive at the following optimization problem:
\begin{align*}
    \Tb_{{\lambda}}^*, \Tb_{{\phi}}^* &:= \arg\min_{\lambda, \phi}~\Lc(\lambda, \phi) \approx \arg\max_{\lambda, \phi} \frac{1}{m}\sum_{i=1}^{m}\log \tilde{p}_{\phi, \lambda}(\zv_i)
\end{align*}
Using Equations \eqref{eq:varQ} and \eqref{eq:pztilde}, we can rewrite our final objective as:
\begin{multline}
\label{eq:obj}
\Tb_{\lambda}^*, \Tb_{\phi}^* =  \arg\max_{\lambda, \phi} \Bigg( \frac{1}{m} \sum_{i=1}^m \log \pi(\thetav_i) + \log p(\varv_i) \\ - \log |\Tb_{\lambda}'(\varv_i ; \thetav_i)| + |\Tb_{\phi}'(\Tb_{\phi}^{-1}\xv_i)|\Bigg)
\end{multline}
where $\forall i \in [m], ~\xv_i = \Tb_{\phi}(\zv_i)$, $\thetav_i = \lfloor \Tb_{\phi}(\zv_i) \rfloor$, $\uv_i = \xv_i - \thetav_i$ and $\varv_i = \Tb_{\lambda}^{-1}(\uv_i)$.

Learning $\Tb_{\lambda}^*, \Tb_{\phi}^*$ results in a density $\tilde{p}^*_{\phi, \lambda}(\zv)$ that is approximately Gaussian with a landscape amenable to efficient MCMC sampling. Thus, our sampling phase consists of running an MCMC sampler of choice with target density $\tilde{p}(\zv)$ resulting in samples $\lbag \zv_1, \zv_2, \cdots, \zv_n \rbag \sim \tilde{p}^*_{\phi, \lambda}(\zv)$. We can finally obtain samples $\lbag \thetav_1, \thetav_2, \cdots, \thetav_n \rbag \sim \pi(\thetav)$ as $\thetav_i = \lfloor \Tb_{\phi}^*(\zv_i) \rfloor$. 

We end this section with two important remarks.

\begin{remark}
In our method, it is possible to by-pass the last step of sampling in the z-space using MCMC entirely since the trained flow is a generative model that can be used directly to generate samples from $\pi(\thetav)$. This can indeed be done if the learned transformations $\Tb_{\lambda}(\cdot ; \thetav)^*$ and $\Tb_{\phi}^*$ result in the density $\tilde{p}(\zv)$ to be Gaussian. Thus, there is a natural trade-off here: we can spend only enough computation to train the flow to learn $\tilde{p}(\zv)$ that is suitable for fast-mixing of the MCMC chain and generate samples from $\tilde{p}(\zv)$ that is not Gaussian using any sampler of choice or we can spend a larger amount of compute to learn a flow that perfectly Gaussianizes the target density $\pi(\thetav)$. Then, we can sample directly by sampling from a Gaussian in the z-space and using the learned transformations to obtain samples from $\pi(\thetav)$.
\end{remark}

\begin{remark}
As mentioned in \Cref{sec:bkg}, we can use any density $p(\zv)$ instead of a Gaussian for training the transformations $\Tb_{\lambda}(\cdot ; \thetav)$ and $\Tb_{\phi}$. The main motivation of our method is to ``push-forward'' $\pi(\thetav)$ onto a space that is amenable to efficient sampling. An interesting future work might be to devise learning objectives that explicitly drive the learned density to have simple geometry that favours ``off-the-shelf'' samplers.
\end{remark}

%% file: sections/prev.tex
\section{Connection to previous works}
\label{sec:prev}
In \Cref{sec:method} we introduced a normalizing flow augmented MCMC sampler for combinatorial spaces. Our method combines surjective transformation for learning continuous relaxations for discrete spaces and normalizing flows that map the  continuous space to a simpler discrete space for easy MCMC sampling. In this section, we put both these ideas of continuous relaxations of discrete spaces and neural transport methods for efficient MCMC sampling in to perspective with existing work.
\vspace{-0.5mm}
\paragraph{Neural transport samplers for continuous spaces:} As we discussed briefly in \Cref{sec:bkg}, the neural transport augmented sampler for continuous variables has been successfully used by \cite{ParnoMarzouk18}, \cite{PeherstorferMarzouk18}, \cite{hoffman2019neutra} and \cite{albergo2019flow}. A subtle difference between these works and our work here -- apart from the major difference that the aforementioned works are applicable only for continuous domains -- is the method of training the transport map itself. These methods train the transport map (or normalizing flow) by minimizing the Kullback-Liebler divergence between the target density $\pi(\thetav)$ and the density $\tilde{\pi}(\thetav)$ learned via $\Tb$ by pushing-forward a standard normal distribution. \cite{albergo2019flow} additionally use the reverse KL-divergence for the target density and the approximate density for training instead.  We, on the other hand, learn by minimizing the KL-divergence in the latent space since we only have access to a lower-bound of the discrete density $\tilde{\pi}(\thetav)$ (\cf \Cref{eq:discpi}).

\paragraph{Continuous relaxations of discrete spaces:} The idea to relax the constraint that random variables of interest only take discrete values has been used extensively in combinatorial optimization \citep{pardalos1996continuous}. %inspired by linear program relaxation. 
Such an approach is attractive since the continuous space affords the function with gradient information, contours, and curvatures that can better inform optimization algorithms. Surprisingly though, this idea has not received much attention in the MCMC setting. Perhaps the closest work to ours in the present manuscript is that of \cite{zhang2012continuous} who use the Gaussian Integral Trick \citep{hubbard1959calculation} to transform discrete variable undirected models into fully continuous systems where they perform HMC for inference and the evaluation of the normalization constant. The Gaussian Integral Trick used by \cite{zhang2012continuous} can be viewed as specifying a fixed map from the discrete space to an augmented continuous space. However, this can result in a continuous space that is highly multi-modal and not amenable for efficient sampling. 

\cite{nishimura2020discontinuous} on the other hand map to a continuous space using \emph{uniform dequantization} \citep{uria2013rnade} \ie filling the space between points in the discrete space with uniform noise inducing parameters with piecewise constant densities. \footnote{This corresponds to $q(\uv|\thetav) = \mathsf{Uniform}(0,1)$ in our method.} They further propose a Laplace distribution for the momentum variables in dealing with discontinuous targets and argue this to be more effective than the Gaussian distribution. This work relies on the key theoretical insight of \cite{pakman2013auxiliary} that Hamiltonian dynamics with a discontinuous potential energy function can be integrated explicitly near the discontinuity such that it preserves the total energy. \cite{pakman2013auxiliary} used this to propose a sampler for binary distributions which was later extended to handle more general discontinuities by \cite{afshar2015reflection}. \cite{dinh2017probabilistic} also used this idea on settings where the parameter space involves phylogenetic trees. A major limitation of \cite{pakman2013auxiliary}, \cite{afshar2015reflection} and \cite{dinh2017probabilistic} is the fact that these method do not represent a general-purpose solver since they encounter computational issues when dealing with complex discontinuities. Similarly, the work of \cite{nishimura2020discontinuous} requires an integrator that works component-wise and is prohibitively slow for high dimensional problems. Furthermore, it is not clear if the the Laplace momentum based sampler leads to efficient exploration of the continuous space which is highly multi-modal due to uniform dequantization.

\paragraph{Discrete MCMC samplers:} A main limitation of embedding discrete spaces into continuous ones is that they can often destroy natural topological properties of the space under consideration \eg space of trees, partitions, permutations etc. \cite{titsias2017hamming} proposed an alternative approach called Hamming ball sampler based on informed proposals that are obtained by augmenting the discrete space with auxiliary variables and performing Gibbs sampling in this augmented space. However, potentially strong correlations between auxiliary variables and the chain state severely slows down convergence. \cite{zanella2020informed} tried to address this problem by introducing locally balanced proposals that incorporate local information about the target distribution. This framework was later called a \emph{Zanella process} by \cite{power2019accelerated}. They used the insights in \citep{zanella2020informed} to build efficient, continuous-time, non-reversible algorithms by exploiting the structure of the underlying space through symmetries and group-theoretic notions. This helps them to build locally informed proposals for improved exploration of the target space. However, their method requires explicit knowledge of the underlying properties of the target space which is encoded in the sampler which can be problematic. 

%% file: sections/exp.tex
\section{Experiments}
\label{sec:exp}
We now present experimental results for our SurVAE Flow augmented MCMC on problems covering a range of discrete models applied in statistics, physics, and machine learning. These include a synthetic example of a discretized Gaussian Mixture Model, Ising model for denoising corrupted MNIST, quantized logistic regression on four real world datasets, and Bayesian variable selection for high-dimensional data. The code for our implementation as well as the experiments is available at \url{https://github.com/priyankjaini/discFlowMH}.
% ranging from 3 dimensional data to 684 dimensions, Bayesian variable selection for high-dimensional data, and sampling from a lattice Gaussian which is prominently used in cryptography.

We compare our model to two baselines that include a random walk Metropolis-Hastings algorithm and Gibbs sampling. We further also compare to discrete HMC (dHMC) \citep{nishimura2020discontinuous} although we use the original implementation released by the authors which is not parallelizable and implements using the numpy package \citep{harris2020array} and R. In contrast, we implemented our method and the other baselines in Pytorch \citep{NEURIPS2019_9015} and is parallelizable to use multiple GPUs for fast and efficient computation. 

For each experiment, we train our model to learn $\Tb_{\lambda}$ and $\Tb_{\phi}$ for $10,000$  iterations 
% in the set $\mathsf{iters} = \{10^3, 2\times10^3, 5\times10^3, 10\times10^3\}$ 
with a batch-size of 128, learning rate of $10^{-3}$ and optimize using the Adam optimizer. We run 128 parallel chains of Metropolis-Hastings algorithm for $10^5$ steps with thinning of 10 (\ie saving every tenth sample) with no burn-in since our learned flow already provides a good initialization for sampling. For fair comparison, we follow a similar setup for the baselines \ie we run 128 parallel chains for each method for $10^5$ steps with thinning of 10. We use $10^5$ burn-in steps for the baselines in all the experiments. Additionally, for denoising binarized MNIST using the Ising model, we also experiment with different number of burn-in steps which are detailed in the results.
% Thus, we experiment with burn-in period given by the set $\mathsf{burn-in}=\{10^4, ~2\times10^4,~ 5\times10^4, ~10\times10^4\}$. 
For dHMC \citep{nishimura2020discontinuous}, we used the publicly available code released by the authors. 

We compare the efficiencies for all these models by reporting both the mean effective sample size (ESS)\footnote{We chose the mean ESS as a metric because for our experiments the minimum ESS across features tended to be close to one due to variables being discrete.} across dimensions, mean ESS across dimensions per minute, and the accuracy or (unnormalized) log-probability of the samples generated by each sampler for the corresponding downstream task. For our method (labeled as Flow + MH), we include the training time of the flow \ie $(\Tb_{\lambda}, \Tb_{\phi})$ when evaluating ESS per min to ensure a fair comparison. We compute the ESS for 16 chains at a time, resulting in 8 ESS estimates and report mean $\pm$ standard error across these 8 estimates.
% Since ESS is fundamentally a metric for univariate samples and all our applications have multivariate parameters, we evaluate the ESS for each dimension separately and report the minimum ESS across dimensions in our results averaged across the chains. 
The major aims of our experimental evaluation presented here are two-fold: Firstly, our aim is to demonstrate the efficacy of our flow augmented MCMC over broad applications. In particular, we want to demonstrate that flow-based generative models present an attractive methodology to sample from a discrete distribution by modelling the distribution in a continuous space and sampling in  a simpler latent space. Thus, in our experiments here we have restricted ourselves to using the basic Metropolis-Hastings (MH) algorithm for sampling to demonstrate the advantages of being able to learn a transport map from the discrete space to a simple latent space. By using a more sophisticated sampler like HMC \citep{duane1987hybrid, neal2011mcmc}, we could thus get even better results.
% Therefore, the experimental results here are a lower bound on the performance of our method since it can be improved with use of advanced samplers like HMC \citep{duane1987hybrid, neal2011mcmc} that have generally better performance compared to MH algorithm. 
Secondly, most samplers (and especially discrete space samplers) are either inefficient and/or are prohibitively slow for high-dimensional problems. Thus, through our experiments we want demonstrate that leveraging the advances of flows to learn high-dimensional distributions, our sampler is more efficient with significantly better results than the alternative methods for high-dimensional problems.
\paragraph{Discretized Gaussian Mixture Model} We first illustrate our method on a 2D toy problem. We define a set of target distributions by discretizing Gaussian mixture models with 5 components using a discretization level of 6 bits. In Fig. \ref{fig:toy} we compare samples from the target distributions, samples from the approximation learned by the flow and samples from the MCMC chain. We observe that the flow gives an initial rough approximation to the target density -- thus greatly simplifying the geometry for the MCMC method. Next, the samples from the MCMC chain are indistinguishable from true samples from the target distribution. This also highlights the trade-off we described in Remark 1 in \Cref{sec:method}.

\newcommand{\fw}{0.31\linewidth}
\begin{figure}[t]
    \centering
\begin{subfigure}[b]{\fw}
    \caption*{Target}
    \vspace{-2mm}
    \includegraphics[width=\textwidth]{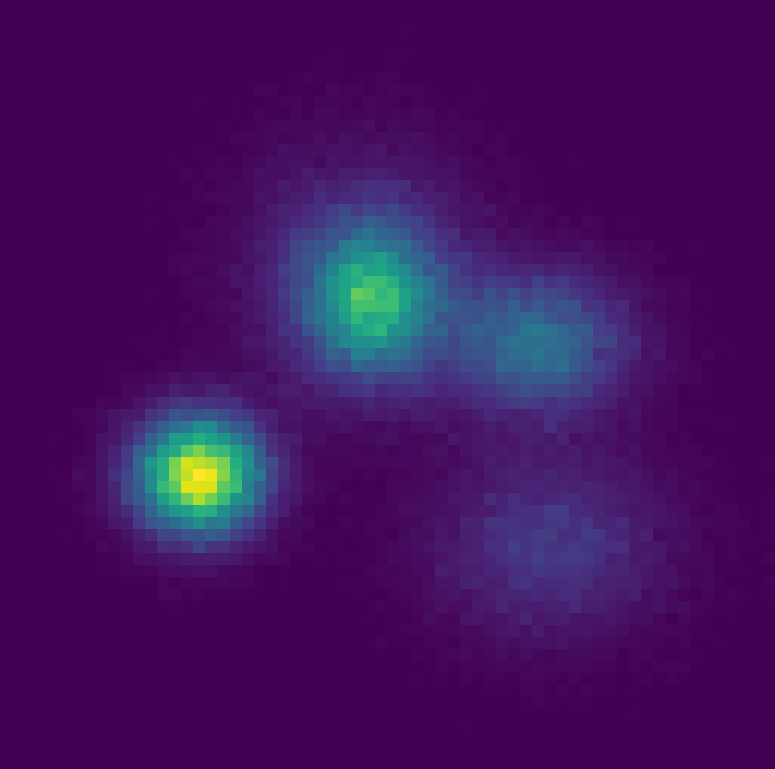}
\end{subfigure}
\begin{subfigure}[b]{\fw}
    \caption*{Flow}
    \vspace{-2mm}
    \includegraphics[width=\textwidth]{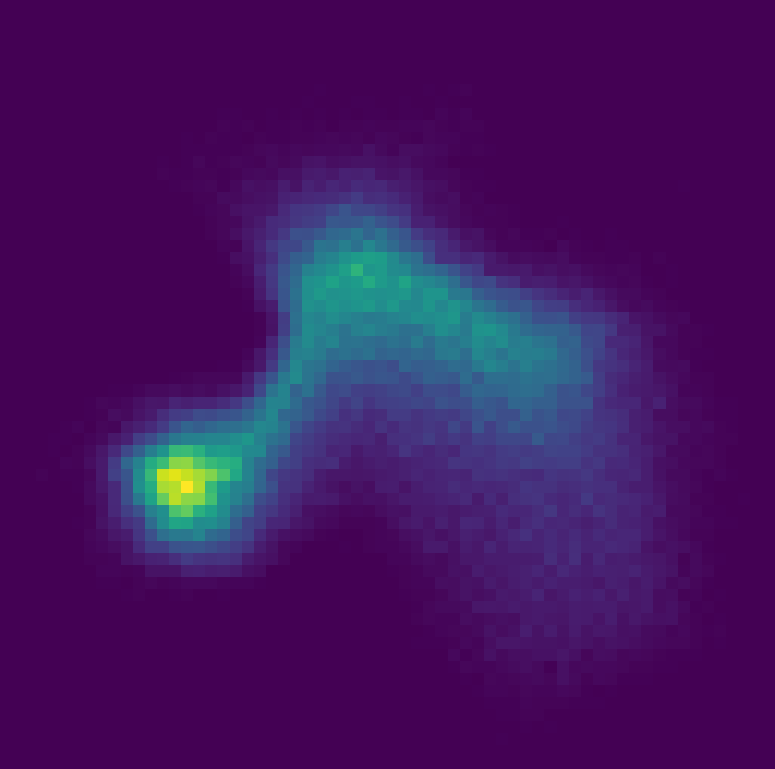}
\end{subfigure}
\begin{subfigure}[b]{\fw}
    \caption*{Flow+MH}
    \vspace{-2mm}
    \includegraphics[width=\textwidth]{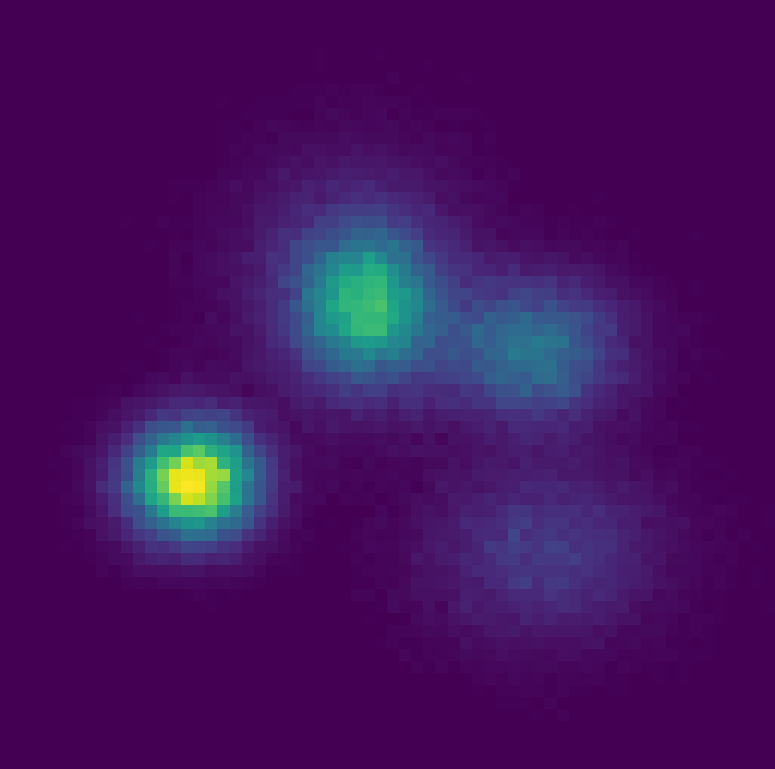}
\end{subfigure}

\begin{subfigure}[b]{\fw}
    \includegraphics[width=\textwidth]{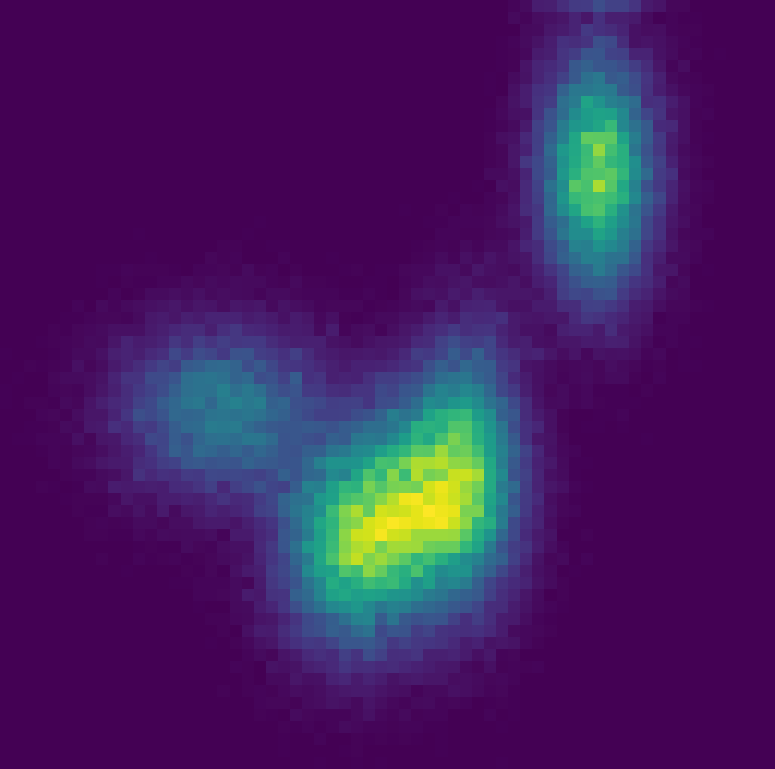}
\end{subfigure}
\begin{subfigure}[b]{\fw}
    \includegraphics[width=\textwidth]{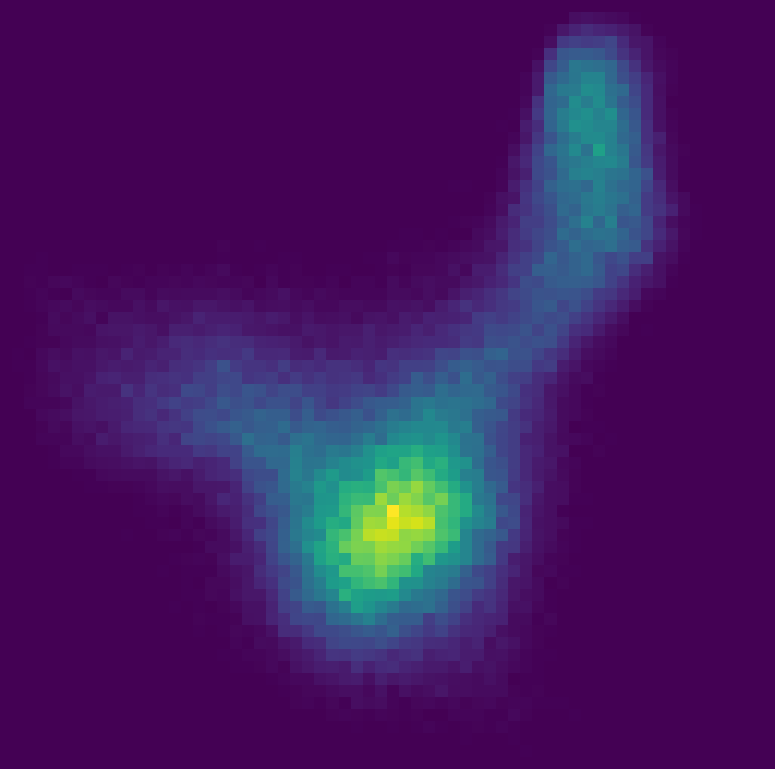}
\end{subfigure}
\begin{subfigure}[b]{\fw}
    \includegraphics[width=\textwidth]{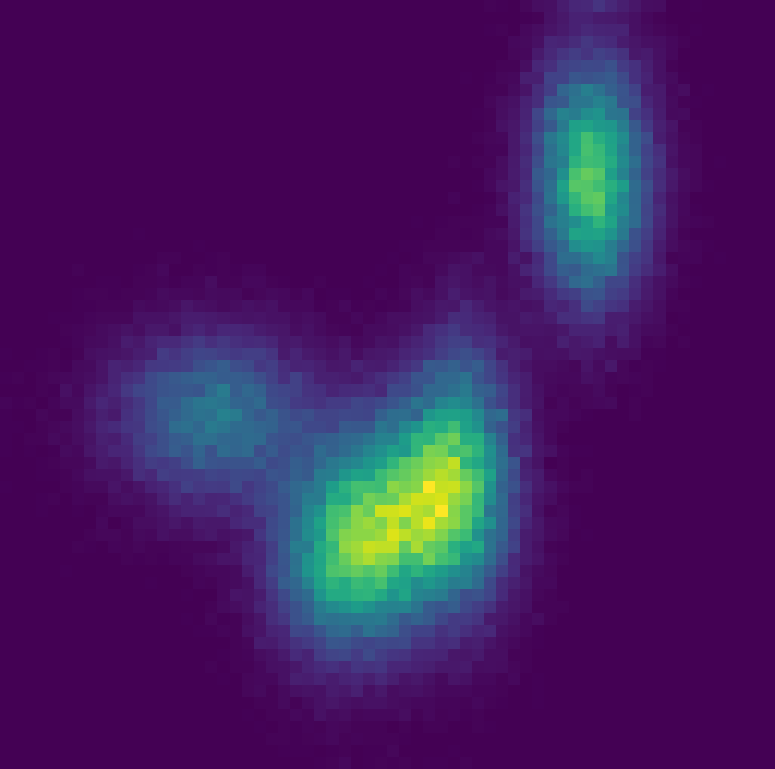}
\end{subfigure}

\begin{subfigure}[b]{\fw}
    \includegraphics[width=\textwidth]{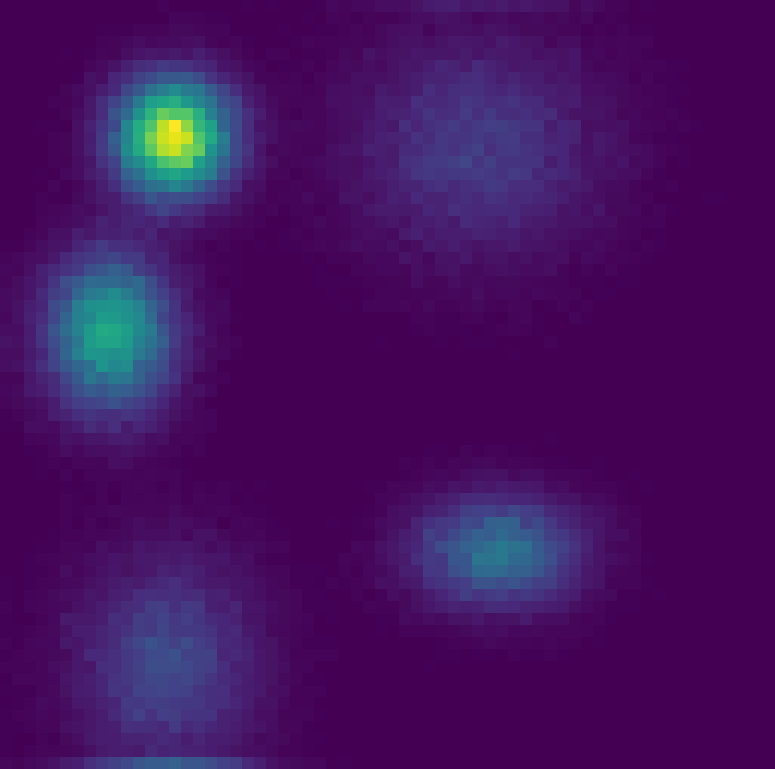}
\end{subfigure}
\begin{subfigure}[b]{\fw}
    \includegraphics[width=\textwidth]{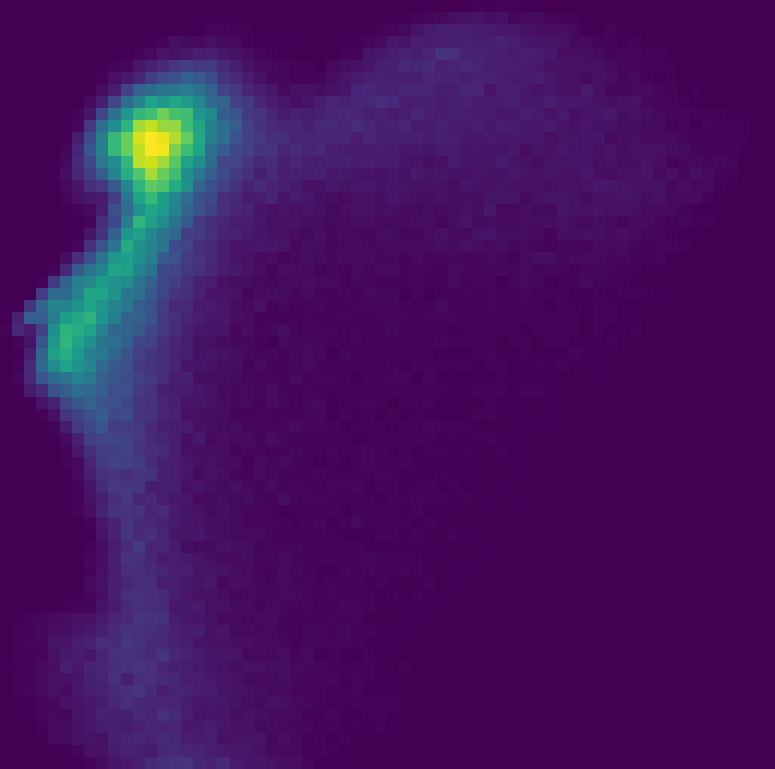}
\end{subfigure}
\begin{subfigure}[b]{\fw}
    \includegraphics[width=\textwidth]{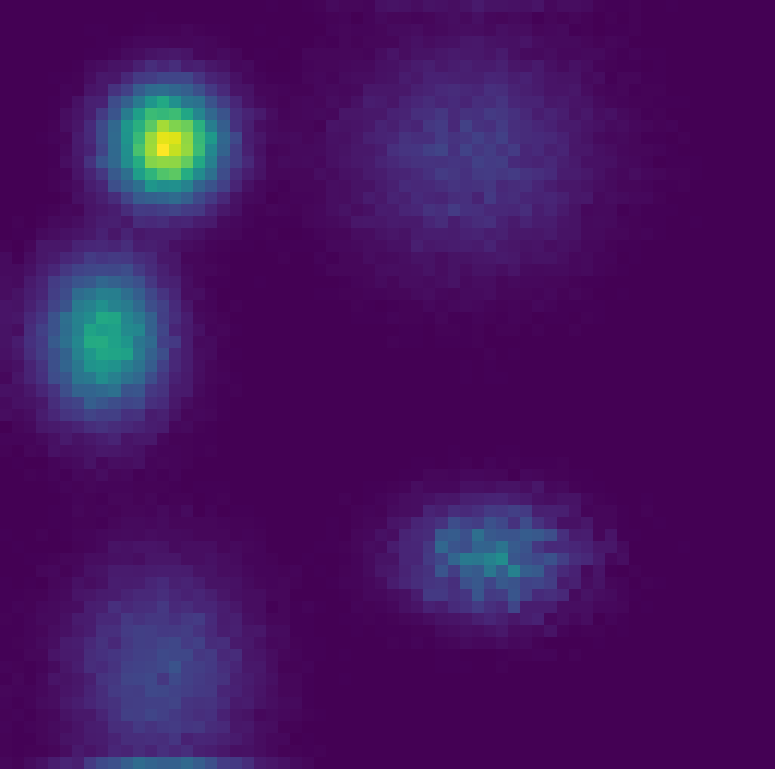}
\end{subfigure}

\begin{subfigure}[b]{\fw}
    \includegraphics[width=\textwidth]{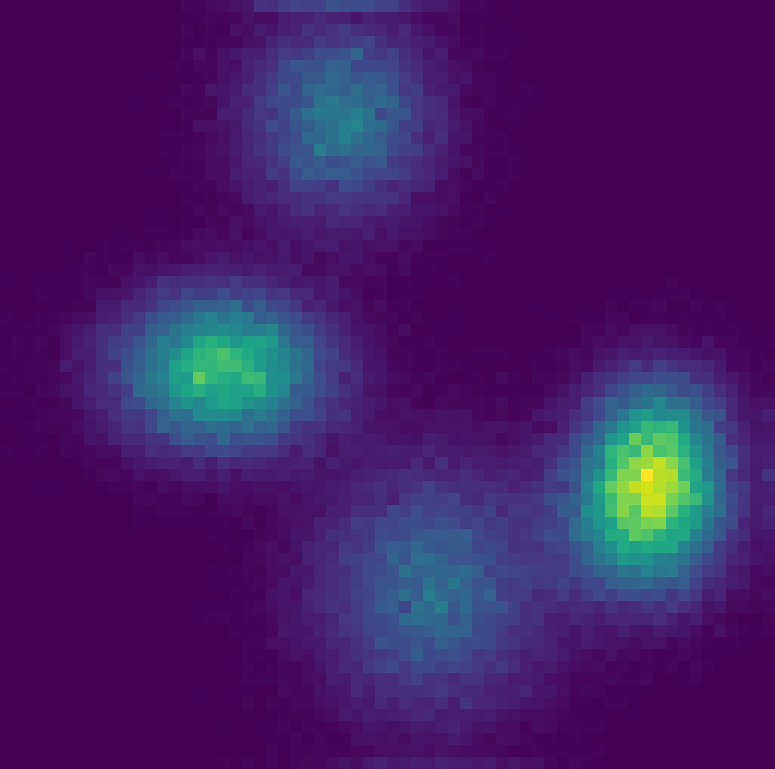}
\end{subfigure}
\begin{subfigure}[b]{\fw}
    \includegraphics[width=\textwidth]{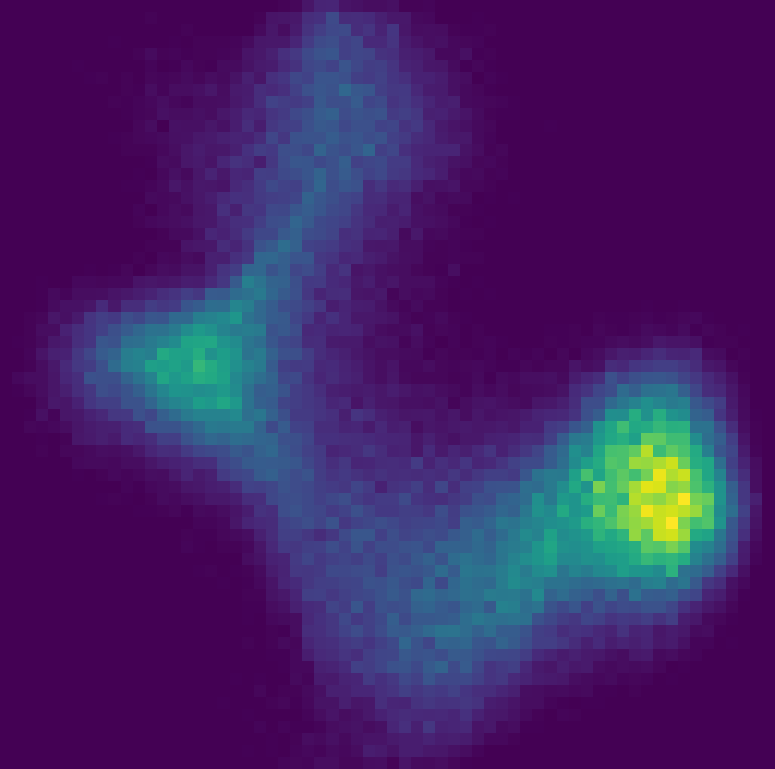}
\end{subfigure}
\begin{subfigure}[b]{\fw}
    \includegraphics[width=\textwidth]{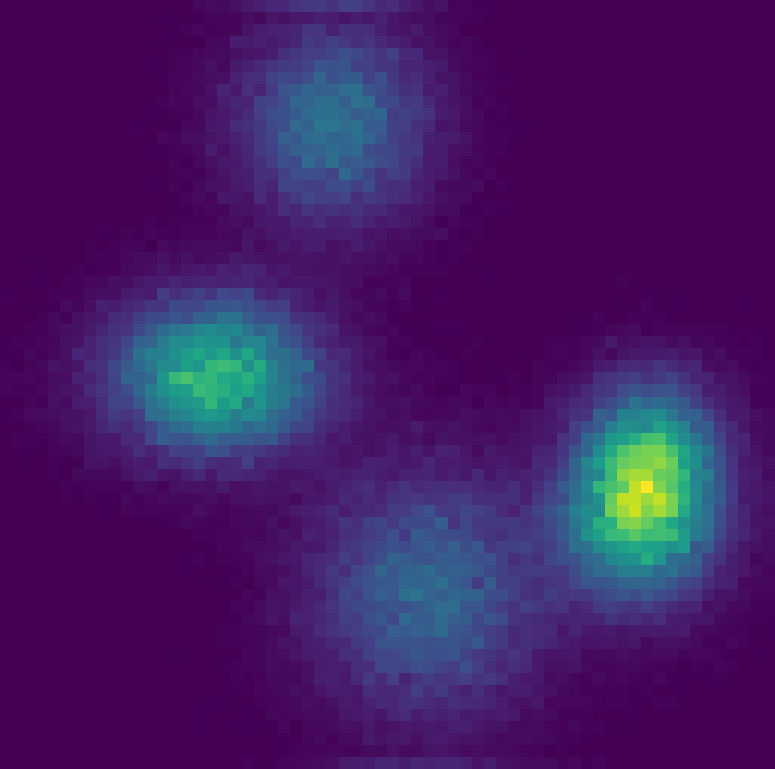}
\end{subfigure}

    \caption{Toy problem. \textit{Left:} Discretized Gaussian mixture target distributions. \textit{Middle:} Approximation learned by flow. \textit{Right:} Samples from MCMC chain.}
    \label{fig:toy}
\end{figure}

\begin{figure}[t]
    \centering
\begin{minipage}[b]{0.31\linewidth}
\begin{subfigure}[b]{\textwidth}
    \caption*{True:}
    \vspace{-2mm}
    \includegraphics[width=\textwidth]{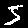}
\end{subfigure}

%  \vspace{4mm}

\begin{subfigure}[b]{\textwidth}
    \caption*{Corrupted:}
    \vspace{-2mm}
    \includegraphics[width=\textwidth]{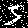}
\end{subfigure}
\end{minipage}
\begin{minipage}[b]{0.53\linewidth}
\begin{subfigure}[b]{5.5cm}
    \centering
    \caption*{Flow+MH:}
    \vspace{-2mm}
    \includegraphics[trim=62 182 0 0,clip,width=0.85\textwidth]{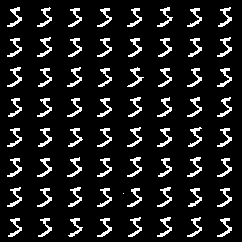}
\end{subfigure}

\begin{subfigure}[b]{5.5cm}
    \centering
    \caption*{Discrete MCMC:}
    \vspace{-2mm}
    \includegraphics[trim=62 182 0 0,clip,width=0.85\textwidth]{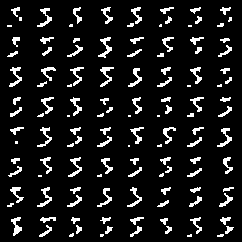}
\end{subfigure}

\begin{subfigure}[b]{5.5cm}
    \centering
    \caption*{Gibbs:}
    \vspace{-2mm}
    \includegraphics[trim=62 182 0 0,clip,width=0.85\textwidth]{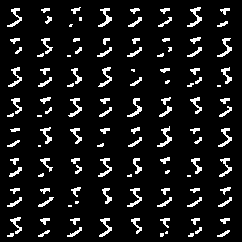}
\end{subfigure}
\end{minipage}

    \caption{Ising model. \textit{Left:} True MNIST digit and observed corrupted MNIST digit. \textit{Right:} Denoised posterior samples from independent MCMC chains.}
    \label{fig:ising}
\end{figure}

\paragraph{Ising Model for Denoising Binarized MNIST}
We illustrate the application of discrete samplers to un-directed graphs in this section by considering the removal of noise from a binary $28 \times 28$ MNIST image. We take an image from the MNIST dataset and binarize it such that the true un-corrupted image is described by a vector $\yv \in \{0, 1\}^{784}$. We obtain a corrupted image $\xv$ by taking this unknown noise-free image $\yv$ and randomly flipping the value of each pixel with probability 0.1. Given this corrupted image, our goal is to recover the original noise-free image.
\begin{table*}[h]
\caption{Results for denoising binary MNIST using the Ising model. We report ESS per $10^4$ samples, ESS per minute, and the average log-likelihood (unnormalized) for the generated samples.}
\vspace{1em}
\centering
%\resizebox{\columnwidth}{!}{
\begin{tabular}{c|c|c|c}
\toprule
Sampler& ESS & ESS/min & $\log \pi(\thetav)$ \\
\midrule
Flow + MH (iters = 1k) & 43.45 $\pm$ 5.07 &  0.15 $\pm$ 0.02 & 4025.67 $\pm$ 7.56\\
 \midrule
Flow + MH (iters = 2k) & \textbf{46.48 $\pm$ 7.02} &  \textbf{0.15 $\pm$ 0.02} & 4039.92 $\pm$ 6.05\\
 \midrule
Flow + MH (iters = 5k) & 45.95 $\pm$ 5.34 & 0.13 $\pm$ 0.01 & 4054.25 $\pm$ 5.15 \\
 \midrule
Flow + MH (iters = 10k) & 28.52 $\pm$ 4.17  & 0.07 $\pm$ 0.01 & 4056.82 $\pm$ 5.64\\ 
\toprule
Gibbs (burn-in = 10k) & 7.131 $\pm$ 1.42 & 0.011 $\pm$ 0.002 & 4045 $\pm$ 22.79 \\
 \midrule
Gibbs (burn-in = 20k) & 6.09 $\pm$ 0.82 & 0.009 $\pm$ 0.001 & 4050.08 $\pm$ 16.56  \\
 \midrule
Gibbs (burn-in = 50k) & 7.65 $\pm$ 0.88  & 0.012 $\pm$ 0.001 & 4056.37 $\pm$ 9.94\\
 \midrule
Gibbs (burn-in = 100k) & 7.82 $\pm$ 1.06& 0.009 $\pm$ 0.001  & 4060.09 $\pm$ 7.56\\
 \toprule
%discrete MH (burn-in = 10k) \\
 %\midrule
discrete MH (burn-in = 10k) & 26.24 $\pm$ 1.84  & 0.07 $\pm$ 0.01 & 3989.26 $\pm$99.77\\
 \midrule
discrete MH (burn-in = 20k) & 12.01 $\pm$ 2.39  & 0.03 $\pm$ 0.01 & 4019.20 $\pm$ 44.22\\
 \midrule
discrete MH (burn-in = 50k) &1.90 $\pm$ 0.17 &  0.004 $\pm$ 0.001 & 4038.96 $\pm$17.95 \\
 \midrule
discrete MH (burn-in = 100k) &2.40 $\pm$ 0.87 & 0.005 $\pm$ 0.002 & 4044.16 $\pm$ 16.03\\
 \bottomrule
 dHMC & 17.83 $\pm$ 1.49 & 0.022 $\pm$ 0.001  & 4038.68 $\pm$ 23.55\\
\bottomrule
\end{tabular}
%}
\label{tab:ising}
\end{table*}
Following \citep[Section 8.3]{bishop2006pattern}, we solve this by formulating it as an Ising model and using Monte carlo samplers to sample $\thetav$ with energy function given by:
\begin{align*}
    E(\thetav, \xv) = -\beta\cdot\sum_{i, j} \thetav_i\thetav_j - \eta\cdot \sum_{i}\thetav_i\xv_i
\end{align*}
where $\thetav_i$ and $\xv_i$ are the $i^{th}$ pixel in the sample under consideration $\thetav$ and corrupted image $\xv$ respectively. 
We train the flow for $\mathsf{iters} = \{10^3, 2\times10^3, 5\times10^3, 10\times10^3\}$ and run the baselines with $\mathsf{burn-in}=\{10^4, ~2\times10^4,~ 5\times10^4, ~10\times10^4\}$.
We report the results in \Cref{tab:ising} and the samples in \Cref{fig:ising}-\ref{fig:ising_more}. We The results evidently show that Flow augmented MCMC significantly outperforms other samplers on both ESS and the underlying downstream task demonstrating its ability to disentangle complex correlations present in high-dimensional problems by mapping it to a latent space whereas other coordinate-wise samplers are not able to handle this efficiently. 

% \begin{figure*}[h]
%     \centering
%     \includegraphics[height=12cm, keepaspectratio]{plots/Ising.png}
%     \caption{Results for Ising Model}
%     \label{fig:ising}
% \end{figure*}

\paragraph{Quantized Logistic Regression}
Next, we consider the task of logistic regression where the learnable parameters and biases denoted by $\thetav$ taking discrete values. This problem is particularly applicable for training quantized neural networks where the weights take discrete values. However, here we restrict ourselves to a simpler version of quantized logistic regression on four real-world datasets from the UCI repository that include Iris, Breast Cancer, Wine, and Digits. For each dataset, we run 5-fold cross-validation -- with results averaged across the 5 folds. In each fold, we train the flow for $10^4$ iterations and run discrete MCMC with a burn-in of $10^5$ steps. We consider parameters quantized to 4 bits and report the results for these in \Cref{tab:qlr}. We notice that our method outperforms other methods on ESS but suffers in ESS/min due to the time it takes to train the flow.

\begin{table*}[!h]
\caption{Results for quantized logisitic regression on four datasets. We report the ESS per $10^4$ samples, ESS per minute, maximum accuracy, and the average log-likelihood for samples generated by each model.}
\vspace{1em}
\centering
\begin{tabular}{c|c|c|c|c|c|c}
\toprule
Dataset & Parameters & Sampler & ESS & ESS/min & accuracy & $\log \pi (\thetav)$ \\
\midrule
\multirow{3}{*}{Iris} & \multirow{3}{*}{15} & Flow + MH & \textbf{923.21 $\pm$ 37.29} & \textbf{58.64 $\pm$ 2.37}  & 97.4 & -8.55 $\pm$ 1.77  \\
\cmidrule{3-7}
& & discrete MCMC & 108.53 $\pm$ 1.44 & 26.14 $\pm$ 0.35 & 97.4 & -8.54 $\pm$ 1.77 \\
\cmidrule{3-7}
& & dHMC & 361.07 $\pm$ 16.97 & 32.85 $\pm$ 0.64 & 97.4 & -8.40 $\pm$ 1.89\\
\midrule
\multirow{3}{*}{Wine} & \multirow{3}{*}{42} & Flow + MH & \textbf{146.25 $\pm$ 3.51} & 6.59 $\pm$ 0.16 & 97.8 & -2.15 $\pm$ 1.63  \\
\cmidrule{3-7}
& & discrete MCMC & 65.78 $\pm$ 0.46 & 15.05 $\pm$ 0.10 & 97.8 & -2.15 $\pm$ 1.63  \\
\cmidrule{3-7}
& & dHMC & 89.12 $\pm$ 11.57 & 4.53 $\pm$ 0.98 & 97.7 & -3.54 $\pm$ 0.48\\
\midrule
\multirow{3}{*}{Breast Cancer} & \multirow{3}{*}{62} & Flow + MH & \textbf{15.77 $\pm$ 0.49} & \textbf{0.56 $\pm$ 0.02} & 96.8 & -19.85 $\pm$ 3.40  \\
\cmidrule{3-7}
& & discrete MCMC & 10.46 $\pm$ 0.12 & 2.29 $\pm$ 0.03 & 96.8 & -20.14 $\pm$ 3.46   \\
\cmidrule{3-7}
& & dHMC & 9.54 $\pm$ 0.28 & 1.38 $\pm$ 0.03 & 96.8 & -21.19 $\pm$ 2.86\\
% \midrule
% \multirow{3}{*}{Digits} & \multirow{3}{*}{650} & Flow + MH &  - & - & 97.4 & -  \\
% \cmidrule{3-7} 
% & & discrete MCMC & 1.51 $\pm$ 0.01 & 0.024 $\pm$ 0.001 & 96.4 & -24.50$\pm$ 5.76 \\
% \cmidrule{3-7}
% & & dHMC &  & 0.54 $\pm$ 0.01 & 95.9 & -22.72 $\pm$ 3.86\\
\bottomrule
\end{tabular}
\label{tab:qlr}
\end{table*}

\paragraph{Bayesian Variable Selection}
\begin{table*}[!h]
\caption{Results for Bayesian variable selection on synthetic datasets of 100, 200, and 400 dimensions with only 10, 20, and 40 informative features, respectively. We report ESS per $10^4$ samples, ESS per minute, and the average log-likelihood for samples generated by each model.}
\vspace{1em}
\centering
\begin{tabular}{c|c|c|c|c}
\toprule
Setting & Sampler & ESS & ESS/min & $\log \pi(\thetav)$\\
\midrule
\multirow{2}{*}{$\mathsf{features}=100$, $\mathsf{features}_{\mathsf{informative}}=10$}& Flow + MH & \textbf{436.48 $\pm$ 57.82} & \textbf{4.78 $\pm$ 0.63} & \textbf{-1054.97 $\pm$ 15.89}\\
\cmidrule{2-5}
  & Gibbs &  23.19 $\pm$ 22.20 & 0.12 $\pm$ 0.12 & -1235.02 $\pm$ 13.11  \\
\toprule
\multirow{2}{*}{$\mathsf{features}=200$, $\mathsf{features}_{\mathsf{informative}}=20$}& Flow + MH & \textbf{213.36 $\pm$ 29.37} & \textbf{0.89 $\pm$ 0.12}  & \textbf{-2231.24 $\pm$ 8.28}  \\
\cmidrule{2-5}
  & Gibbs & 8.78 $\pm$ 6.12 & 0.013 $\pm$ 0.009 & -2485.91 $\pm$ 0.14  \\
\toprule
\multirow{2}{*}{$\mathsf{features}=400$, $\mathsf{features}_{\mathsf{informative}}=40$}& Flow + MH & \textbf{141.45 $\pm$ 17.97} & \textbf{0.15 $\pm$ 0.02} &\textbf{ -4627.58 $\pm$ 10.86}\\
\cmidrule{2-5}
  & Gibbs &  1.46 $\pm$ 0.23 & 0.001 $\pm$ $10^{-4}$ & -5306.61 $\pm$ 0.20 \\

\bottomrule
\end{tabular}
\label{tab:br}
\end{table*}
Here, we consider the problem of probabilistic selection of features in regression where the hierarchical framework allows for complex interactions making posterior exploration extremely difficult. Following \cite{schafer2013sequential}, we consider a hierarchical Bayesian model in which the target vector $\yv \in \mathbb{Z}^d$ in linear regression is observed through:
\begin{align*}
    \yv|\beta, \thetav, \sigma^2, \xv \sim \Nc(\xv\cdot\thetav \cdot \beta, \sigma^2 I)
\end{align*}
with features $\xv \in \RR^{d \times k}$, parameters $\beta \in \RR^k$ and the binary vector $\thetav$ that indicates the features that are included in the model. To achieve a closed form for the marginal posterior that is independent of $\beta$ and $\sigma^2$, we make the following choices following \cite{power2019accelerated}:
$\pi(\thetav | \xv, \yv) = \int \pi(\thetav|\xv, \yv, \beta, \sigma^2) \rmd \beta ~\rmd \sigma^2$
with conjugates $p(\beta | \sigma^2, \thetav) = \Nc(0, \nu^2\sigma^2 I_n \thetav)$, $p(\sigma) = \text{I} \Gamma(\frac{2}{2}, \frac{\alpha w}{2})$ and, $p(\thetav) = \mathsf{Uniform}(\{0, 1\}^k)$ where $\text{I} \Gamma$ is the inverse-gamma distribution and we set $\nu, w, \alpha$ as in \cite{george1997approaches}. For this setting, we create high-dimensional synthetic datasets rather than the low-dimensional datasets usually used to demonstrate the efficacy of our method. We present the results in \Cref{tab:br}.

%% file: sections/con.tex
\section{Conclusion}
\label{sec:con}
In this paper, we presented a flow based Monte Carlo sampler for sampling in combinatorial spaces. Our method learns a deterministic transport map from a discrete space to a simple continuous latent space where it is efficient to sample. Thereby, we sample from the discrete space by generating samples in the latent space and using the transport map to obtain samples in the discrete space. By learning a map to a simple latent space (like standard Gaussian), our method is particularly suited for high-dimensional domains where alternative samplers are not efficient due to the presence of more complex correlations. This is also reflected in our implementation which is faster and efficient as demonstrated by our results on a suite of experiments. In the future, it will be interesting to devise learning strategies for the transport map that explicitly pushes the latent space to have certain desirable properties for efficient sampling. Another direction could be to extend the framework of SurVAE Flow layers to incorporate underlying symmetries and invariance in the target domain.

%% file: sections/appendix.tex
\section{Additional Results}

\begin{figure}[h!]
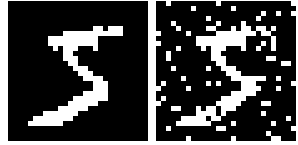
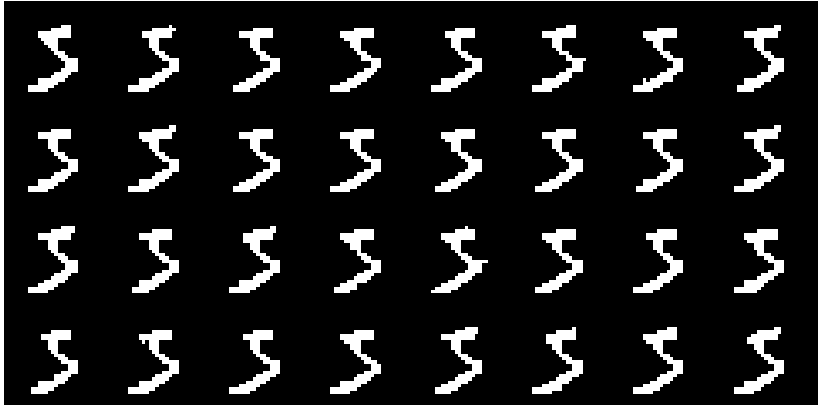
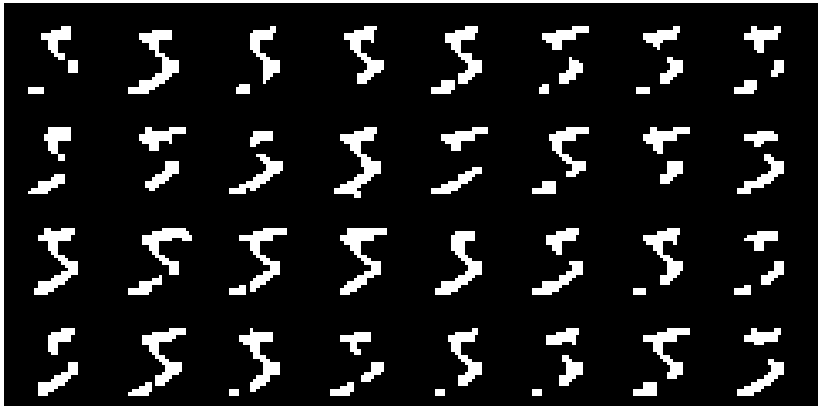
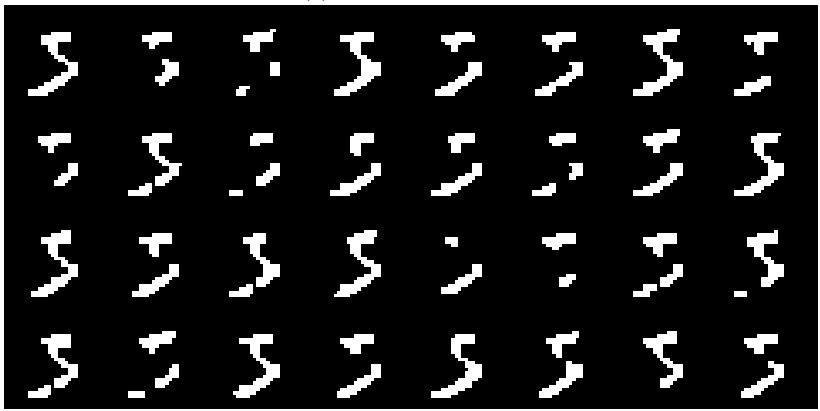

    \centering
    
\begin{subfigure}[t]{0.49\textwidth}
    \centering
    \includegraphics[width=0.22\textwidth]{plots/ising/data_true.png}
    \includegraphics[width=0.22\textwidth]{plots/ising/data_corr.png}
    \caption{True MNIST digit and observed corrupted MNIST digit.}
\end{subfigure}

\begin{subfigure}[t]{0.7\textwidth}
    \centering
    \includegraphics[trim=0 122 0 0,clip,width=0.9\textwidth]{plots/ising/flow.png}
    \caption{Flow+MH.}
\end{subfigure}

\begin{subfigure}[b]{0.7\textwidth}
    \centering
    \includegraphics[trim=0 122 0 0,clip,width=0.9\textwidth]{plots/ising/dmcmc.png}
    \caption{Discrete MCMC.}
\end{subfigure}

\begin{subfigure}[b]{0.7\textwidth}
    \centering
    \includegraphics[trim=0 122 0 0,clip,width=0.9\textwidth]{plots/ising/gibbs.png}
    \caption{Gibbs.}
\end{subfigure}

    \caption{Original and corrupted MNIST digit together with denoising posterior samples from different MCMC methods. Samples are from independent MCMC chains.}
    \label{fig:ising_more}
\end{figure}

We here provide additional results for the Ising model experiment in Sec. \ref{sec:exp}. The results are shown in Fig. \ref{fig:ising_more}.

% \begin{figure}[h!]
%     \centering
    
% \begin{subfigure}[t]{0.49\textwidth}
%     \centering
%     \includegraphics[width=0.45\textwidth]{plots/ising/data_true.png}
%     \includegraphics[width=0.45\textwidth]{plots/ising/data_corr.png}
%     \caption{Original (left) and corrupted (right) image.}
% \end{subfigure}
% \begin{subfigure}[t]{0.49\textwidth}
%     \centering
%     \includegraphics[trim=0 122 0 0,clip,width=0.9\textwidth]{plots/ising/flow.png}
%     \caption{Flow+MH.}
% \end{subfigure}

% \begin{subfigure}[b]{0.49\textwidth}
%     \centering
%     \includegraphics[width=0.9\textwidth]{plots/ising/dmcmc.png}
%     \caption{Discrete MCMC.}
% \end{subfigure}
% \begin{subfigure}[b]{0.49\textwidth}
%     \centering
%     \includegraphics[width=0.9\textwidth]{plots/ising/gibbs.png}
%     \caption{Gibbs.}
% \end{subfigure}

%     \caption{Ising model. \textit{Left:} True MNIST digit and observed corrupted MNIST digit. \textit{Right:} Denoised posterior samples from independent MCMC chains.}
%     \label{fig:ising_more}
% \end{figure}